\documentclass{article}
\usepackage{microtype}
\usepackage{graphicx}
\usepackage{subfigure}
\usepackage{booktabs}
\usepackage{hyperref}

\usepackage[accepted]{icml2023_deployableGenAI}
\usepackage{amsmath}
\usepackage{amssymb}
\usepackage{mathtools}
\usepackage{amsthm}

\usepackage[capitalize,noabbrev]{cleveref}

\theoremstyle{plain}

\theoremstyle{definition}

\theoremstyle{remark}

\usepackage[textsize=tiny]{todonotes}

\def\eqn#1{\begin{equation}#1\end{equation}}

\begin{document}

\twocolumn[
\icmltitle{Discovering Variable Binding Circuitry with Desiderata}

\icmlsetsymbol{equal}{*}

\begin{icmlauthorlist}
\icmlauthor{Xander Davies}{equal,harvard}
\icmlauthor{Max Nadeau}{equal,harvard}
\icmlauthor{Nikhil Prakash}{equal,northeastern}
\icmlauthor{Tamar Rott Shaham}{mit}
\icmlauthor{David Bau}{northeastern}
\end{icmlauthorlist}

\icmlaffiliation{harvard}{Harvard University}
\icmlaffiliation{northeastern}{Northeastern University}
\icmlaffiliation{mit}{MIT CSAIL}

\icmlcorrespondingauthor{Xander Davies}{xanderlaserdavies@gmail.com}
\icmlcorrespondingauthor{Max Nadeau}{mnadeau@college.harvard.edu}

\icmlkeywords{Mechanistic Interpretability, Variable Binding, Automatic Circuit Discovery, Language Model}

\vskip 0.3in
]



\printAffiliationsAndNotice{\icmlEqualContribution} 



\begin{abstract}
    Recent work has shown that computation in language models may be human-understandable, with successful efforts to localize and intervene on both single-unit features and input-output circuits. Here, we introduce an approach which extends causal mediation experiments to automatically identify model components responsible for performing a specific subtask by solely specifying a set of \textit{desiderata}, or causal attributes of the model components executing that subtask. As a proof of concept, we apply our method to automatically discover shared \textit{variable binding circuitry} in LLaMA-13B, which retrieves variable values for multiple arithmetic tasks. Our method successfully localizes variable binding to only 9 attention heads (of the 1.6k) and one MLP in the final token's residual stream.
\end{abstract}

\section{Introduction} 

Deploying powerful generative AI systems requires confidence in the reliability of their outputs, especially with respect to certain high stakes behaviors like manipulation or truthfulness \citep{carroll2023characterizing, perez2022discovering}. The emerging field of \textit{mechanistic interpretability} seeks to make model computation human-understandable by explaining the function of particular model components and locating groups of model components responsible for performing certain language tasks. Indeed, recent work has successfully identify, localize and intervene 
in model computation \citep{othello22, burns2022discovering, wang2022interpretability, conmy2023automated}.

Here, we introduce an automated approach which extends activation patching \citep{rome, vig2020investigating} to localize components within neural networks (\emph{e.g.} attention heads, MLP layers) that responsible for performing a specific subtask of model computation. Our method allows for quickly and automatically localizing computation, while only requiring to specify \textit{desiderata}, or causal attributes of the target computation. As a proof of concept, we apply our method to automatically discover shared \textit{variable binding circuitry} in LLaMA-13B \citep{llama23}, which retrieves variable values for several arithmetic operations. 

\begin{figure*}[t]
    \centering
    \includegraphics[width=1\linewidth]{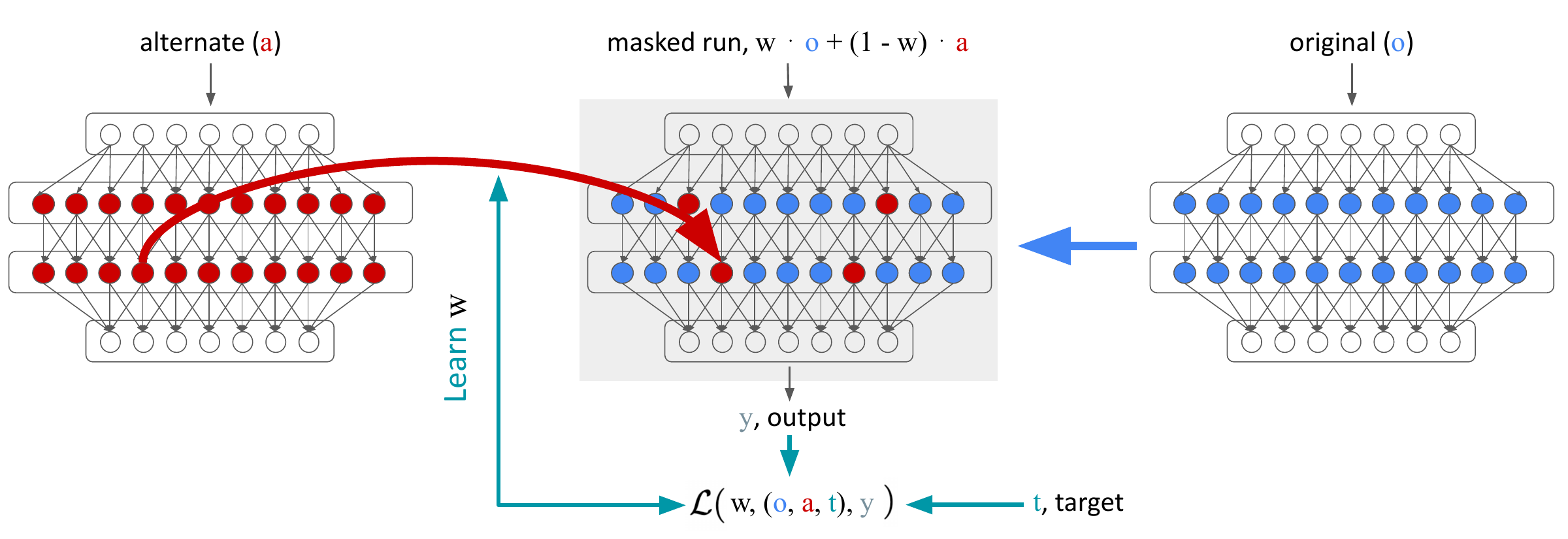}
    \caption{\textbf{Localizing computation with desiderata.} The figure depicts training with a single (original, alternate, target) tuple within a desideratum. We learn a mask $w$ that combines activations from an alternate sequence $a$ into the computation of the model on the input of the original sequence $o$ such that the output $y$ moves towards the target $t$.}
    \label{fig:subcirc}
\end{figure*}

\paragraph{Contributions.} In this ongoing work, we:
\begin{enumerate}
    \item Describe a methodology for localizing computation by enumerating desiderata and learning a binary mask by performing causal interventions (Section~\ref{sec:desiderata}, Fig.~\ref{fig:subcirc}).
    \item Present initial results in applying this methodology to localize shared \textit{variable binding} circuitry (Section~\ref{sec:variable-binding}, Fig.~\ref{fig:vb-desir}).
\end{enumerate}

\section{Background} 

\paragraph{Circuit analysis.}  A deep neural network can be represented as a directed acyclic graph with specific nodes to accept inputs, generate outputs, and perform various operations to transform inputs into outputs. Circuit analysis involves localizing and understanding subgraphs within the computational graph of a model that are responsible for specific behaviors, and has had success in both language and vision models \citep{olah2020zoom, wang2022interpretability, raukur2022toward, chan_causal_2022}. 

\paragraph{Activation Patching.} As introduced in \citep{rome}, activation patching is a technique that uses causal intervention to identify which submodules' activations matter for producing some model output. The process of activation patching involves running all the layers of a model until reaching a certain submodule with an original input, denoted as $A$, and a corrupted input, denoted as $B$. The activations of this specific submodule with input $B$ are then patched into the corresponding activations of the same submodule with input $A$ during the forward pass. Next, the patched activations are fed forward through the rest of the model. This enables one to assess the role of the specific submodule in generating an output, by quantifying how much this intervention shifts the model's output from its original answer on $A$. Our approach generalizes activation patching and swaps activation from multiple alternative input sequence runs with known target outputs, instead of corrupted sequence runs, as described in \citep{rome}. 



\paragraph{Variable Binding.}
Variable binding is the process of associating a variable with a specific value, and is a fundamental concept in symbolic reasoning considered essential for solving tasks such as natural language understanding and reasoning \citep{gary2001}. However, it is still a mystery if and how Large Language Models (LLMs) 
implement this process.

Please see Appendix~\ref{sec:related-work} for additional related work.

\section{Using Desiderata to Localize Computation}
\label{sec:desiderata}

We discover circuitry responsible for a specific task by enumerating properties of such a desired circuitry, and then learning a binary mask over the model's parameters which accords with these properties (Fig.~\ref{fig:subcirc}). We specify  properties (or desiderata) in terms of causal interventions with a known target effects, and combine various interventions into a single objective function. We then learn a sparse mask on the targeted model components, such that applying causal interventions on the masked components alters model behavior to satisfy the objective function.

\paragraph{Model components.} As a first step, we specify our set of \textit{model components}. Models can be represented at various levels of granularity. More granular components is more computational expensive, but allows for more specific localization of a model behavior. In Section \ref{sec:variable-binding}, we decompose LLaMA-13B into a set of attention heads and MLPs, as opposed to more granular (e.g. splitting by Query, Key, and Value matrices) or less granular (e.g. grouping into layers) representations.



\paragraph{Desiderata.} Given a computational circuitry with specific functionality, we define a set of desiderata to enumerate the effects of various causal interventions on the circuitry. 
Each desideratum $d$ corresponds to a set of $n$ 3-tuple, each of which consists of an original sequence ($o$), an alternate sequence ($a$), and a target value ($t$). 
When the activation of the sought-after circuitry generated with $o$ is replaced with the corresponding activation generated with $a$, the model should output $t$. The target value ($t$) is determined based on the nature of the intervention: it can remain equal to the output of $o$ (indicating no change in the output is expected), be altered to match the output of $a$, or be set to a completely different third value. We identify and localize submodule with the desired functionality 
based on its adherence to the expected outcomes specified by the desiderata.

\begin{figure*}[t]
    \centering
    \includegraphics[width=0.8\linewidth]{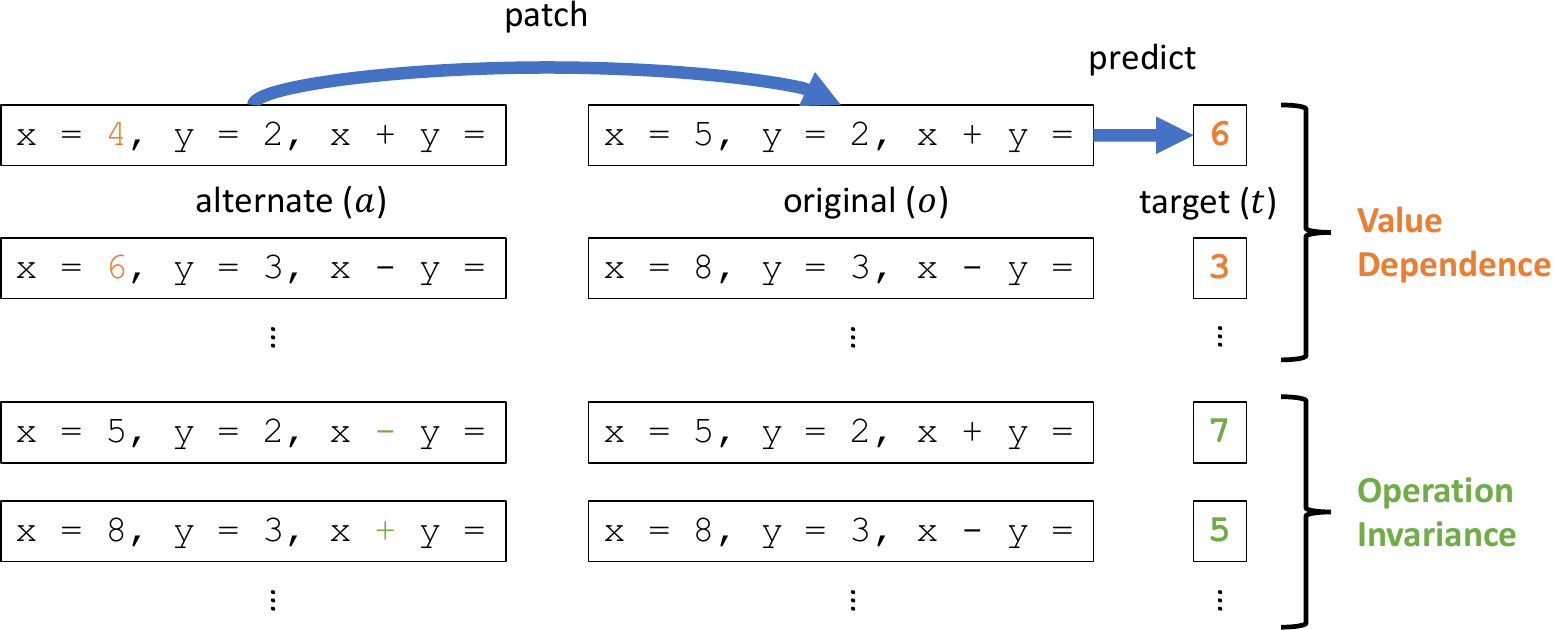}
    \caption{\textbf{Variable Binding Desiderata.} Each desideratum is a set of original ($o$), alternate ($a$), and target ($t$) 3-tuples. In the \textit{Value Dependence} desideratum, patching should change the output to the alternate's output; in the \textit{Operation Invariance} desideratum, patching should have no effect.}
    \label{fig:vb-desir}
\end{figure*}

Each 3-tuple $(o,a,t)$  contributes to a loss term which measures how well performing activation patching on a set of model components $\{c_i\}$ achieves $t$,  \eqn{\label{eqn:desir} \mathcal{L}_d(\{c_i\}) = \frac{1}{n} \sum_{(o, a, t) \in d}\mathcal{L}(\{c_i\}, (o, a, t), y).} Note that some measure of proximity $\mathcal{L}$ between the induced model output $y$ and the target $t$ is needed. 
Furthermore, one can combine multiple desideratum into a single objective function, $\mathcal{L}_{D}(\{c_i\}) = \sum_{d \in D} \mathcal L_d(\{c_i\})$. Desiderata for the specific case of identifying the value-copying circuitry 
involved in variable binding are presented in Fig.~\ref{fig:vb-desir} and discussed in Section~\ref{sec:vb-desir}.

\paragraph{Learning a Binary Mask.}  In order to find the set of model components $\{c_i\}$ that minimizes $\mathcal L_D(\{c_i\})$, we use a continuous relaxation of Equation~\ref{eqn:desir}. We define a mask over the model components by assigning a learnable weight $w_i \in [0, 1]$, to each component $c_i$. That is, $w_i = 0$ corresponds to fully patching component $c_i$ with its value $v_a$ from the sequence $a$, $w_i = 1$ corresponds to not patching $c_i$, and $0 < w_i < 1$ corresponds to taking a convex combination of the $c_i$'s activation value $v$ and the value $v_a$ 

\eqn{w_i \cdot v + (1-w_i)\cdot v_a.}

Note that when we patch multiple components of the model, the value $v$ of a later-layer component will be influenced by the patching of earlier layers before it itself is combined with $v_a$ as above.

We optimize the continuous mask according to $\mathcal L_D$, which measures how well the patching intervention defined by the mask meets our desiderata. We use $\ell_{0.5}$ regularization with tunable strength $\lambda$ over the mask entries to encourage patching only a sparse set of model components  \citep{louizos_learning_2018}. Throughout learning, we clamp values between 0 and 1. After training, we round weights to either 0 or 1 to form a binary mask. Empirically, we find that rounding the mask to become binary typically has little effect on its ability to satisfy the desiderata, and attribute this to the regularization for sparsity during training.

\section{Variable Binding}
\label{sec:variable-binding}

\begin{table*}[t]
\centering
\resizebox{\linewidth}{!}{%
\begin{tabular}{l|rrrrr}
 & \textbf{VD Acc. (+, -)} & \textbf{OI Acc. (+, -)} & \textbf{VD Acc. ($\times$)} & \textbf{OI Acc. (+, $\times$)} & \textbf{\# Patched} \\
\hline
Original Model & 18\% & 91\% & 11\% & 93\% & 0 \\
\hline
Incomplete Desiderata (VD) & 93\% & 11\% & 82\% & 13\% & 10\\
Full Desiderata (VD \& OI) & 84\% & 82\% & 84\% & 91\% & 10\\
\end{tabular}
}
\caption{\textbf{Accuracy of patching experiments.} Learning the patching mask according to an incomplete set of desiderata (only using the Value Dependence desideratum as presented in the second row) fails to localize our target computation (Operation Invariance accuracy suffers for all the tested operation indicated in parenthesis).
Using both desiderata (Full Desiderata. third row) successfully causes Value Dependence behavior while maintaining Operation Invariance. Interestingly, the learnt patching mask achieves high accuracy also for operation that was not included in the training set. 
For all cases, accuracy is calculated on held-out test set.}
\label{tab:results_table}
\end{table*}

We apply our method (Section~\ref{sec:desiderata}) to locate circuitry responsible for retrieving variable values when computing simple arithmetic expressions like those in Fig.~\ref{fig:vb-desir}. We use LLaMA-13B, a 40-layer, decoder-only transformer language model, trained on a diverse data set \citep{llama23}. We hypothesize that there exist components of LLaMA-13B that, in order to complete sequences like those appearing in Fig.~\ref{fig:vb-desir}, copy the value previously assigned to the variable $x$ into the final token's residual stream. We further hypothesize that the $x$'s value is then combined with $y$'s value to compute the desired expression. 

We design desiderata to search specifically for this value-copying circuitry. Throughout, we only evaluate accuracy of models based on whether their prediction of the first digit of the answer value is correct; we ensure a diverse set of targets to avoid degenerate solutions. Code to replicate our results is available in a public repository.\footnote{\url{https://anonymous.4open.science/r/anima_submission-D770/README.md}}

\subsection{Variable Binding Desiderata}
\label{sec:vb-desir}

We propose two desiderata to isolate this hypothesized value-copying circuitry:\footnote{We note that additional desideratum are possible.}
\begin{enumerate}
    \item \textit{Value Dependence} (VD; Fig.~\ref{fig:vb-desir}, top). Patching our target circuitry with its activations from alternate sequences containing different $x$ values should control which value is copied into the final residual stream. Accordingly, such patching 
    should change the model's output to match the output of the alternate sequences.
    \item \textit{Operation Invariance} (OI; Fig.~\ref{fig:vb-desir}, bottom). Since we are looking for circuitry shared across arithmetic operations, the specific operation being performed in the expression should not affect the behavior of the value-copying circuitry, as it should copy the same variable value regardless of the  operation. 
    We therefore form alternate 
    sequences 
    with a flipped operation (either addition or subtraction), with a target value equals to the output value of the corresponding original sequence. 
\end{enumerate}

\subsection{Binary Mask Details}

We consider 
all MLPs and attention heads (1640 models' components in total) and learn a binary mask as described in Section~\ref{sec:desiderata}. We only perform patching to each component's contribution to the final-token residual stream, as that is where we expect the value-copying circuitry to be active. We use two-digit variable values with addition and subtraction operations for defining both VD and OI sequences. We use the logit difference between the original and alternate answers as the proximity measure. For the VD task, we intend to maximize the logit difference, whereas, for the OI task, we aim to minimize it.



We created a dataset comprising VD and OI sequences, with a total of $90/90$ train/test examples, 
such that the first digit of the expected answer is uniformly drawn from $[1, 9]$. We use the Adam optimizer \citep{kingma2017adam} with a learning rate of 0.01, and alternate between taking gradient steps from the VD loss and the OI loss to save memory. For all experiments 
described below we use a sparsity regularization weight of $\lambda=0.03$. 
In Appendix~\ref{sec:reg-strength} we present additional results with different experimental settings such as varying $\lambda$ and the numbers of patched attention heads. 

\section{Results}
\label{sec:results}

According to our desiderata, we have identified a set of ten components, comprising nine attention heads and one MLP, that execute variable binding. 
We patch heads according to this ten-component mask and evaluate the model's accuracy on a held-out set of VD and OI problems. On VD scenarios, accuracy measures how often the model outputs the answer from the alternative sequence. On OI scenarios, accuracy measures how often the model outputs the answer from the original sequence. We expect a mask that finds heads corresponding to the value copying subtask of variable binding to achieve high accuracy in both VD and OI.

We observe that the models' components identified by our method exhibit high accuracy in both tasks, as indicated in Table~\ref{tab:results_table} (first and second columns). We further test these components on VD problems involving a multiplication operation instead of addition or subtraction, and on OI problems involving swapping between addition and multiplication. Surprisingly, we find that the accuracy remains high in these scenarios, despite not including multiplication during training (see Table~\ref{tab:results_table}, third and fourth columns).
This indicates that these ten components indeed serve as the circuitry that copies variable values to the final residual stream (before the model operates on them); these components successfully cause the model's output to change in the case that one of the bound values changes, but not in the case that the operation in the equation changes, even when testing an operation that was not include in training.

We also find that including both desiderata is crucial for locating this circuitry; with only the VD desideratum, the identified heads 
successfully alter model behavior in the VD scenario, but also affect the model's output in the OI scenario (see Table~\ref{tab:results_table}, second row). When both desiderata are included in the loss, the masked components are mostly attention heads in the middle of the model;\footnote{Heads 11.11, 12.0, 12.7, 15.11, 15.25, 17.17, 18.11, 18.18, 19.20, and MLP 27.} 
whereas when only using the first desideratum, the masked components form a cluster of late-layer MLPs\footnote{MLPs 18, 27, 28, 29,  30, 31, 32, 33, 35, and 36.}. A possible explanation for this is that using only the VD desideratum, the mask includes model components that write the computed final value of the expression to the residual stream, whereas adding the second desideratum encourages the mask to find the value-copying circuitry.

\section{Conclusion}
\label{sec:next-steps}

In this paper, we proposed a new approach to localizing model components responsible for performing a specific task, using a set of causal behavior desiderata. Our method localize 10 components responsible for copying variable values in LLaMA-13B. 
We plan to compare it with existing localization methods 
\citep{meng2022locating, conmy2023automated} and to expend it to additional tasks.

\section*{Acknowledgments}
XD and MN would like to thank Sam Marks and Oam Patel for their valuable discussions. NP and DB are supported by grants from Open Philanthropy. TRS is supported in part by the Zuckerman STEM Leadership
Program and the Viterbi Fellowship.

\bibliography{references}  
\bibliographystyle{icml2023}

\newpage
\appendix
\section{Varying regularization strength}
\label{sec:reg-strength}
We vary the regularization strength $\lambda$ in order to learn masks with varying numbers of heads. We find that setting the regularization so that the masks learns approximately 10 model subcomponents is the approximate minimum number that can score highly for held-out Value Dependence accuracy and Operation Invariance accuracy, and so we use that setting for the main results of our paper. We also show further that removing the Operation Invariance desideratum causes the mask to score poorly on that criterion.

\begin{figure}[t]
    \centering
    \includegraphics[width=1\linewidth]{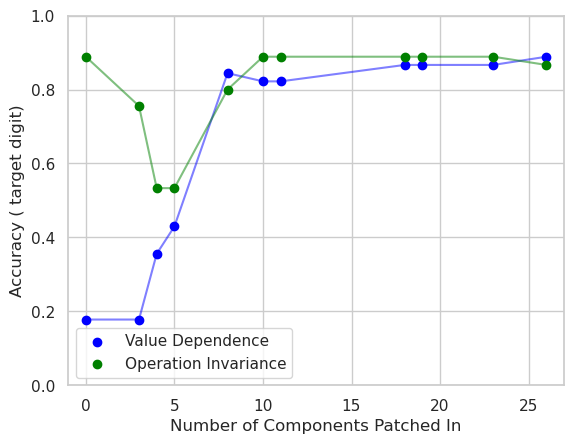}
    \caption{\textbf{Evaluating masks of various numbers of heads on held-out VD and OI problems.} Each vertical pair of datapoints corresponds to a mask learned by a training run with a different value of $\lambda$, the sparsity regularization weight. With too few components patched, the model does not score well at Value Dependence. We interpret this as indicating that not enough of the value-copying heads have been patched.}
    \label{fig:varied-lambda}
\end{figure}

\begin{figure}[t]
    \centering
    \includegraphics[width=1\linewidth]{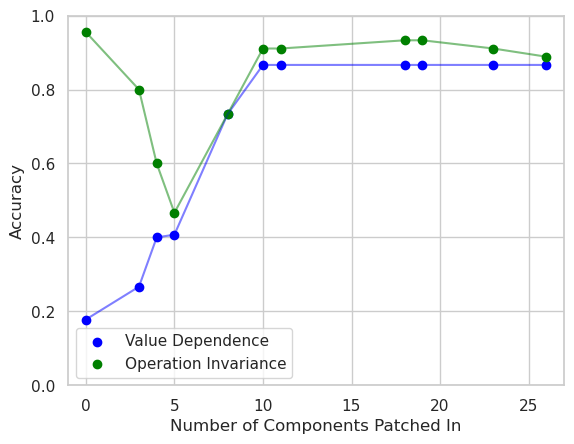}
    \caption{\textbf{Transfer to accuracy on multiplication problems.} This graph depicts the same masks as Fig.~\ref{fig:varied-lambda} (which were trained on sequences involving only addition and subtraction), but evaluated on all-multiplication Value Dependence problems, and addition-to-multiplication (and vice versa) Operation Invariance problems. Similarly to Fig.~\ref{fig:varied-lambda}, VD accuracy is low with too few heads patched.}
\end{figure}

\begin{figure}[t]
    \centering
    \includegraphics[width=1\linewidth]{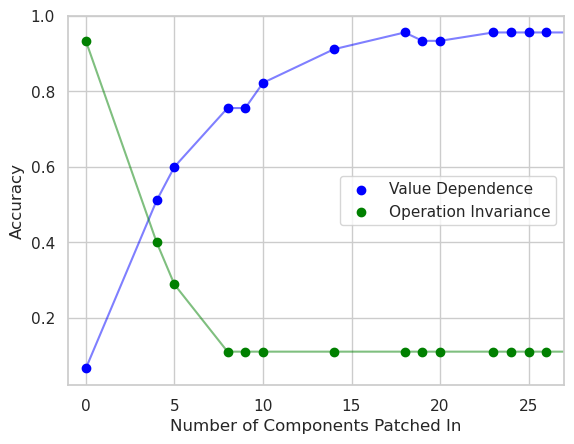}
    \caption{\textbf{Varying regularization strength with incomplete desiderata.} This graph demonstrates learning a mask with only the Value Dependence desideratum. Again, each vertical pair of datapoints corresponds to a mask learned by a training run with a different value of $\lambda$, the sparsity regularization weight. Unlike when the mask is optimized according to both desiderata, these masks fail to achieve high accuracies on both Operation Invariance and Value dependence at the same time, as discussed in Section~\ref{sec:variable-binding}.}
\end{figure}

\section{Related Work}
\label{sec:related-work}

Previous work has developed automated approaches to localizing computation \citep{conmy2023automated, geiger2022inducing, wu2023interpretability}. Our work varies from \citet{conmy2023automated} in learning a mask and considering a broader class of ablations (patches to change behavior, instead of just preserve). Our work shares features with recent work from \citet{geiger2023finding} and \citet{wu2023interpretability}, but differs in attempting to isolate shared computation common in multiple input-output circuits as opposed to understanding full input-output circuits.

\end{document}